\PassOptionsToPackage{table}{xcolor}
\documentclass[]{fairmeta}

\usepackage{amsmath}
\usepackage{amssymb}

\usepackage{multicol}
\usepackage{tabularx}
\usepackage{enumitem}
\usepackage{multirow}
\usepackage[normalem]{ulem}
\usepackage{booktabs}
\usepackage{colortbl}
\usepackage{array}
\usepackage{arydshln}

\usepackage{etoolbox}

\makeatletter
\patchcmd{\mymaketitle}
  {\includegraphics[width=1.5cm]{assets/logometa.pdf}}
  {}
  {}
  {\PackageWarning{main}{Could not remove the Meta logo automatically}}
\makeatother

\title{Computational Humor with Multimodal LLMs:
Methods, Datasets, Evaluation, and Challenges}

\author[1,*]{\href{mailto:tuo.liang@case.edu}{Tuo Liang}}
\author[2,*]{\href{mailto:zhe-derek.hu@connect.polyu.hk}{Zhe Hu}}
\author[1]{\href{mailto:disheng.liu@case.edu}{Disheng Liu}}
\author[2]{\href{mailto:wjc2830@gmail.com}{Juncheng Wang}}
\author[2]{\href{mailto:jing-amelia.li@polyu.edu.hk}{Jing Li}}
\author[1,\dagger]{\href{mailto:yu.yin@case.edu}{Yu Yin}}

\affiliation[1]{Case Western Reserve University}
\affiliation[2]{The Hong Kong Polytechnic University}

\definecolor{headerblue}{RGB}{47,85,151}
\definecolor{rowgray}{RGB}{242,245,249}
\definecolor{accentblue}{RGB}{68,114,196}

\definecolor{groupA}{RGB}{242,245,249}
\definecolor{groupB}{RGB}{255,255,255}
\definecolor{groupC}{RGB}{242,245,249}
\definecolor{groupD}{RGB}{255,255,255}
\definecolor{groupE}{RGB}{242,245,249}

\definecolor{naGray}{RGB}{150,150,150}

\newcolumntype{Y}
  {>{\raggedright\arraybackslash\hsize=.8\hsize}X}

\newcolumntype{Z}
  {>{\raggedright\arraybackslash\hsize=1.3\hsize}X}

\newcolumntype{T}
  {>{\raggedright\arraybackslash\hsize=1.1\hsize}X}

\abstract{
Multimodal humor in memes, cartoons, and comics remains difficult for AI
systems because intended meaning depends on non-literal mechanisms, shared
cultural knowledge, and communicative intent rather than literal scene
description. This survey focuses on visual humor understanding in single-image
and multi-panel artifacts, while treating humor generation as an emerging
downstream frontier. We position the literature against prior humor, sarcasm,
and general MLLM surveys and organize it using a capability-centric hierarchy
spanning recognition, interpretation and reasoning, and generation. Under this
lens, we synthesize benchmark design, evaluation protocols, and modeling
paradigms, tracing the field's shift from task-specific fusion models to
large-model approaches based on multimodal alignment, evidence-grounded
reasoning, and controlled generation. We conclude by highlighting the main
barriers to progress: shortcut-prone evaluation, limited cultural and narrative
coverage, weak evidence grounding, and unresolved safety and ownership
concerns.
}

\begin{document}

\maketitle

\begingroup
\renewcommand{\thefootnote}{\fnsymbol{footnote}}

\footnotetext[1]{These authors contributed equally.}

\footnotetext[2]{Corresponding author.}

\endgroup




\section{Introduction}


Recent advances in AI have enabled models to jointly process text and images with unprecedented scale and performance. Multimodal Large Language Models (MLLMs) now achieve strong results on tasks such as image captioning, visual question answering, and cross-modal retrieval~\cite{yin2024survey,comanici2025gemini,bai2025qwen3vltechnicalreport}, which primarily emphasize the recognition, grounding, and reasoning of explicit visual and textual content.
However, this paradigm remains limited when handling content whose meaning extends beyond literal perception~\cite{hwang2023memecap,chen2024we, nayak2024benchmarking,xu2025visulogic}. Human communication frequently relies on expressive artifacts such as memes, comics, cartoons, and satirical images, where meaning arises from humor, cultural reference, symbolic association, or intentional incongruity between modalities. Interpreting such media requires reasoning about implicit meaning and communicative intent rather than merely recognizing observable content.



We refer to this problem setting as \textbf{multimodal humor understanding}: interpreting image-based artifacts whose humorous, satirical, or ironic meaning emerges from the interaction of visual content and associated text. We focus on single-image and multi-panel image artifacts, and we treat humor generation as an emerging downstream frontier that is useful for probing whether models have internalized the mechanisms of humorous interpretation, rather than as an independent capability to be evaluated in isolation.


\noindent\textbf{Why this gap matters.}
The gap between perceptual recognition and communicative interpretation is not a minor residual error; it is a structural blind spot in how MLLMs are built and evaluated. A model can correctly identify a burning house, a courtroom, or a smiling face, and still fail to recover the metaphor, the social target, or the punchline that those elements jointly encode. This is because the core difficulty is not multimodal alignment in the usual sense, but reasoning about why a juxtaposition is funny, what background knowledge it presupposes, and what stance it communicates to an audience~\cite{saakyan2024v,ryan2025humor}. As MLLMs are increasingly deployed in applications such as content moderation, social media analysis, and creative assistance, systematically understanding their capabilities, limitations, and failure modes in these tasks becomes increasingly urgent.

\smallskip
\noindent\textbf{The gap in existing surveys.} 
Despite growing interest, existing surveys only  cover this space partially. Surveys of text-based computational humor~\cite{amin2020survey, kalloniatis2024computational, loakman2025s, lemmens2026computational} center on linguistic phenomena such as puns, wordplay, and verbal irony, but do not address visual grounding or cross-modal incongruity. Multimodal surveys exist, but each narrows to a single task: sarcasm detection~\cite{farabi2024survey, gao2025spoken} or meme classification~\cite{afridi2020multimodal, ren2026survey}. Broader MLLM surveys~\cite{yin2024survey}, meanwhile, typically subsume creative and figurative understanding as one capability under general multimodal reasoning, without dedicated systematic treatment.

This survey is designed to fill the space. Rather than grouping work by task name or model architecture, we organize the literature around three progressively demanding capabilities: \emph{recognition}, where systems detect or classify humorous phenomena; \emph{interpretation and reasoning}, where systems explain the mechanism, target, or implicit meaning; and \emph{generation}, where models must produce humor-consistent outputs grounded in the input. This framing clarifies not only what different benchmarks measure, but also which modeling assumptions are needed to move from label prediction to evidence-grounded interpretation.

Concretely, this survey makes the following contributions:
\begin{itemize}[leftmargin=*]
    \item We propose a capability-centric organization that aligns background concepts, benchmark design, and modeling paradigms.
    \item We synthesize datasets and evaluation protocols with particular attention to what current benchmarks do and do not measure about interpretive understanding.
    \item We identify the technical and socio-technical bottlenecks that currently limit progress, including shortcut-prone evaluation, sparse mechanism-level annotation, weak cultural grounding, and safety and ownership concerns.
\end{itemize}

\noindent\textbf{Survey scope.}
We include Multimodal Visual Humor, such as \emph{multimodal humor}, \emph{meme understanding}, \emph{visual sarcasm}, \emph{satire detection}, \emph{comic understanding}, \emph{humorous captioning}, and \emph{visual figurative language}. We focus on contemporary MLLMs work, and briefly introduce task-specific multimodal models before the MLLM era in Appendix~\ref{before_mllm} We include papers that study image-based artifacts and where success depends on reasoning beyond literal grounding, typically through visual--text interaction, non-literal mechanisms, or communicative intent. We exclude audio- and video-centric humor, generic aesthetic modeling, and persuasive media unless they are directly used as evaluation resources for image-based humor understanding. We additionally used backward and forward snowballing from benchmark and survey papers to recover closely related work. Our goal is not an exhaustive catalog of every humor-adjacent dataset, but a principled synthesis of the resources and modeling strategies that most directly test multimodal visual humor understanding.

\begin{figure*}[t]
	\centering
	\includegraphics[width=0.98\linewidth]{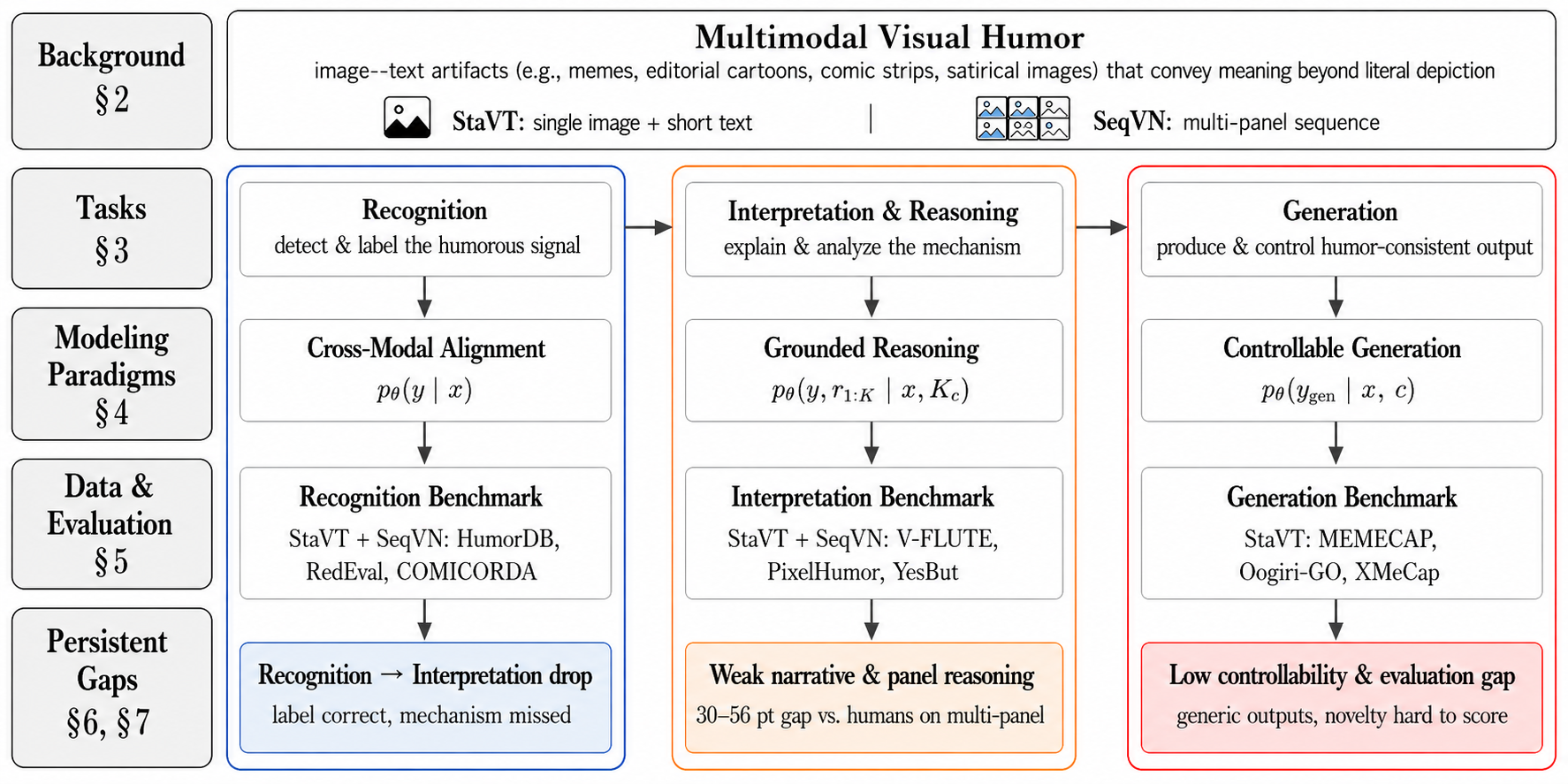}
	\vspace{-0.5em}
	\caption{\textbf{Overview of the survey.} We define \emph{multimodal humor} and its two \emph{representation forms}: StaVT (single image + short text) and SeqVN (multi-panel sequence). Three progressively demanding \emph{capabilities}--- Recognition, Interpretation \& Reasoning, and Generation---form the columns (\S\ref{sec:tasks}). Row~3 gives the dominant \emph{modeling paradigm} for each capability: cross-modal alignment $p_\theta(y \mid x)$, grounded reasoning $p_\theta(y,r_{1:K} \mid x,\mathcal{K}_x)$, and controllable generation $p_\theta(y_{\mathrm{gen}} \mid x,c)$ (\S\ref{sec:modeling}, Table~\ref{tab:paradigm_comparison}). Row~4 lists the \emph{benchmark family and evaluation protocol}, tagged by representation form (\S\ref{sec:datasets}, Table~\ref{tab:dataset_by_capability_level}); Row~5 (accented) highlights the \emph{persistent gaps} exposed by our cross-benchmark analysis (\S\ref{sec:empirical},~\S\ref{sec:challenges}). Horizontal arrows mark capability progression; vertical arrows trace each column top-to-bottom from task to open problem.}
	\vspace{-1.0em}
	\label{fig:tasks_specific}
\end{figure*}

\section{Background}
\label{taxonomy}

\subsection{Definition of Multimodal Humor}

In this survey, we use \textbf{multimodal humor} to refer to communicative artifacts that intentionally convey meaning beyond literal depiction through the coordinated use of visual content and associated text, such as internet memes, editorial cartoons, comic strips, and satirical images.

Unlike natural images or descriptive text, human creative media are designed to communicate expressive, humorous, narrative, or satirical meaning, requiring interpretation of what is implied rather than what is directly shown.
Their interpretation relies not only on perceptual recognition but also on inference of implicit intent and shared social knowledge. We focus on single-image and multi-panel image artifacts where meaning emerges from visual--text interaction and from non-literal reasoning.

\subsection{Representation Forms}

Multimodal humor spans a wide range of image--text data forms, including single images with captions and multi-panel sequences with dialogue. Modeling multimodal creative understanding can be formulated as learning a conditional mapping: $p(y|x)$, where $x$ denotes multimodal inputs and $y$ represents a task-specific output, such as a humor label, explanation, or generated continuation. The structure of $x$ determines the perceptual encoding, alignment mechanisms, and architectural components required to model this conditional distribution.

Unlike conventional vision or language tasks, creative media often distribute meaning across modalities and (for comics) across panels. Therefore, representation form defines not only the input space but also the model’s required integration capacity. We categorize representation forms into two modeling-oriented classes:
\begin{figure*}
    \centering
    \includegraphics[width=1\linewidth]{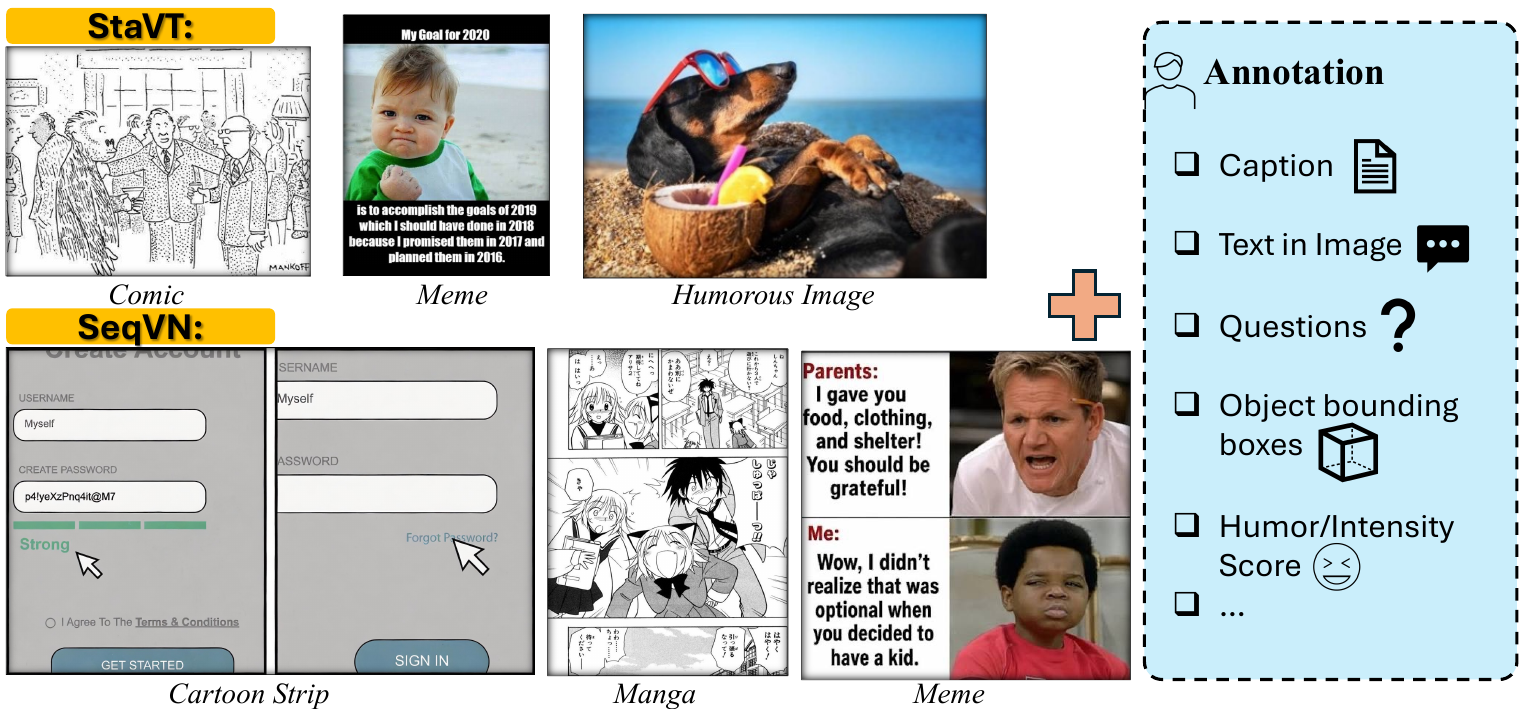}
    \caption{We categorize visual humor data into Static Visual–Textual Artifacts (StaVT), such as comics~\cite{hessel2023androids}, memes, and humorous images~\cite{jain2025humordb}, and Sequential Visual Narratives (SeqVN), such as cartoon strips~\cite{liang2025yes}, manga~\cite{ikuta2025mangaub}, and multi-panel memes. Common annotations include captions, image text, questions, object bounding boxes, humor or intensity scores, and task-specific labels.}
    \label{fig:placeholder}
\end{figure*}


\smallskip
\noindent\textbf{(1) Static Visual–Textual Artifacts (StaVT)} include memes, editorial cartoons, and captioned images, where $x=\{x^{img}, x^{text}\}$ consists of a single image optionally paired with short text. Meaning frequently arises from cross-modal incongruity or implicit symbolic alignment. Modeling requires joint embedding spaces, multimodal alignment, and sensitivity to social knowledge priors~\cite{shifman2013memes,sharma2020semeval}. Architectures typically rely on vision–language encoders or MLLMs with cross-attention mechanisms.


\smallskip
\noindent\textbf{(2) Sequential Visual Narratives (SeqVN)} encompass comic strips and multi-panel memes, where meaning emerges from progression across panels, with $x=\{x_1, x_2,\ldots,x_T\}$ representing an ordered visual sequence.
Understanding these forms requires tracking entities and events, inferring causal relations, and integrating information across a visual sequence~\cite{paval-etal-2025-comicscene154,wang-etal-2025-beyond-single}.

\smallskip
Across all categories, data form shapes not only perceptual requirements but also the types of reasoning and alignment needed for understanding and generating creative meaning.

\subsection{Creative Meaning Construction}

Beyond representation form, creative artifacts convey meaning through non-literal mechanisms. For AI systems, interpretation therefore requires more than recognizing visual and textual patterns: models must also infer why those patterns are used~\cite{hu2023language}. In this survey, we treat creative meaning construction as the interaction between recurrent expressive mechanisms.

Across data forms, four mechanisms recur. \textbf{Incongruity} creates meaning through a mismatch between expectation and observation, often through cross-modal conflict in multimodal humor~\cite{forabosco1992cognitive,veale2004incongruity,schifanella2016detecting,farabi2024survey}. \textbf{Analogy and Conceptual Mapping} project structure from a familiar source domain onto a target concept, as in metaphor and symbolic representation~\cite{lakoff2024metaphors,refaie2003understanding,foss2004theory}. \textbf{Exaggeration and Hyperbole} amplify attributes against implicit norms for emphasis or affect~\cite{kreuz1996figurative,zhang2024image}, while \textbf{Narrative Structure} organizes setup, payoff, and causal progression over time~\cite{genette1980narrative,bruner1991narrative,paval-etal-2025-comicscene154}. Together, these mechanisms explain how creative artifacts encode meaning beyond surface semantics and why they remain challenging for literalist AI systems.

\subsection{Why Multimodal Humor Is Hard for AI}

Multimodal humor is fundamentally challenging for AI models because the intended meaning often cannot be directly inferred from observable inputs. Unlike literal multimodal tasks where answers are grounded in explicit visual or textual evidence, creative understanding requires reasoning over latent variables such as implied metaphors, violated expectations, cultural references, and communicative intent. Models must detect incongruity, infer implicit norms, integrate external socio-cultural knowledge, and reason about why an artifact was produced and how it is meant to be interpreted. As a result, creative understanding goes beyond multimodal feature fusion, demanding the integration of perceptual alignment, structured reasoning, and pragmatic inference within a unified modeling framework.

\begin{figure*}[t]
	\centering
	\includegraphics[width=0.99\linewidth]{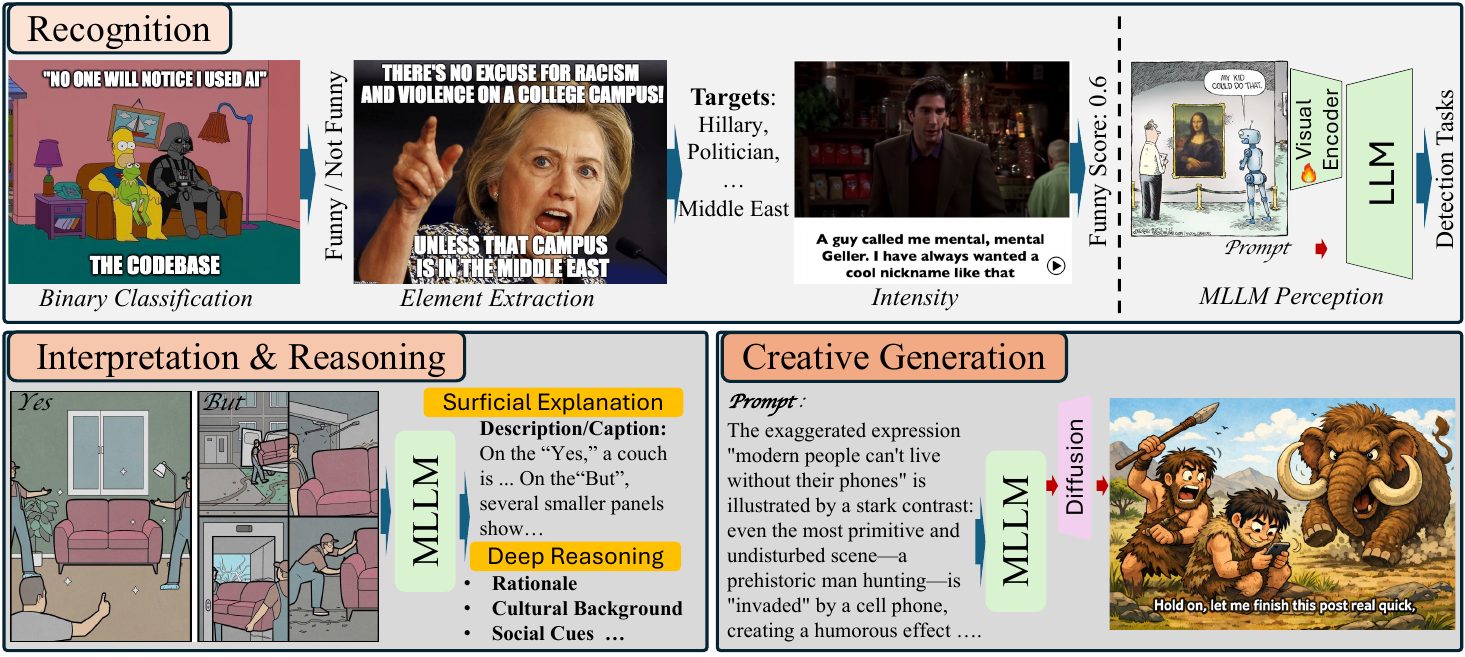}
	\vspace{-0.5em}
	\caption{Capability-centric task hierarchy for multimodal humor. We organize tasks into three levels—Recognition, Interpretation and Reasoning, and Generation—reflecting increasing requirements for non-literal grounding, mechanism-aware inference, and evaluation beyond discriminative labels.}
	\vspace{-1.0em}
	\label{fig:tasks_specific}
\end{figure*}

\section{Task Hierarchy: Recognition, Interpretation, and Generation}
\label{sec:tasks}
We organize existing work into three capability levels and pair each with the evaluation signal it most directly demands (Figure~\ref{fig:tasks_specific}). The hierarchy is not purely chronological: recent benchmarks often mix levels, but the distinction is useful because gains in label prediction do not automatically transfer to explanation or generation.

\subsection{Level 1: Recognition}
Recognition tasks ask whether a multimodal input contains a humorous, sarcastic effect and, in more fine-grained settings, which components instantiate it. Typical problems include binary or multi-class classification, target or role extraction, and intensity estimation. These tasks dominate early work on multimodal sarcasm, meme classification, and visual humor detection~\cite{cai2019multi,hasan2021humor,zhang2024image}. They require reliable visual-text alignment and sensitivity to local cues, but they do not by themselves verify whether a model has captured the mechanism that makes an artifact funny or critical.

Recognition tasks are usually evaluated with discriminative metrics such as accuracy, F1, AUROC, or correlation with human ratings. These metrics are appropriate for testing cue sensitivity and class separation, yet they remain weak proxies for genuine understanding: a model can predict a correct label by exploiting recurrent surface patterns without being able to explain the intended meaning.

\subsection{Level 2: Interpretation and Reasoning}
Interpretation tasks ask models to explain \emph{why} an artifact is humorous, satirical, or ironic. A useful distinction is between \textbf{descriptive explanation}, which verbalizes salient cues or paraphrases the joke, and \textbf{mechanism-grounded reasoning}, which identifies the specific conflict, analogy, target, or narrative step that produces the effect~\cite{hu2024cracking,saakyan2024v,wang2024mementos}. These tasks require explicit cross-modal grounding, abstraction over non-literal meaning, and often external knowledge or context to resolve implicit references. For multi-panel inputs, they also require temporal and narrative reasoning across panels rather than within a single image~\cite{paval-etal-2025-comicscene154,wang-etal-2025-beyond-single}.

Evaluation at this level must combine language quality with interpretive faithfulness. Automatic metrics such as BLEU or BERTScore~\cite{papineni2002bleu,zhang2019bertscore} are useful for checking lexical or semantic overlap, but they remain insufficient when several explanations are plausible. Stronger protocols ask whether a model identifies the relevant cues, names the right mechanism, and stays consistent with available evidence, often through rubric-based human judgment or LLM-assisted evaluation~\cite{liu2023g,hu2024cracking}.

\subsection{Level 3: Generation}
Generation tasks require models to produce humor-consistent outputs such as captions, punchlines, explanations, or comic continuations conditioned on an input artifact. In this survey, we treat generation as an emerging downstream frontier rather than the core of the field: successful generation presupposes at least partial understanding of incongruity, target selection, tone, and narrative setup~\cite{hwang2023memecap,li2023oxfordtvg,tanaka-etal-2024-content}. The challenge is therefore not only to produce fluent text, but also to maintain faithfulness to the source image and control over the mechanism being realized.

Because valid outputs are diverse, evaluation at this level relies heavily on human preference judgments or rubric-based assessment. Reference-based metrics remain weak proxies for humor quality and novelty, so the most informative benchmarks combine generation quality with tests of faithfulness to the source image, intended target, and rhetorical device.

\section{Modeling Paradigms in the Large-Model Era}
\label{sec:modeling}



With the rise of MLLMs, multimodal humor modeling has shifted from handcrafted fusion toward alignment-driven representation learning, explicit reasoning, and evidence-grounded generation. We organize the modeling literature by the \textbf{capability} it primarily supports rather than by model family, because the same backbone behaves very differently when optimized for \textit{recognition}, \textit{interpretation}, or \textit{controlled generation}. Table~\ref{tab:paradigm_comparison} summarizes the three resulting paradigms along their input formulation, technical core, and characteristic failure modes; the strongest recent systems combine them rather than treating them as substitutes. We detail each paradigm below and close with a cross-paradigm analysis.

\begin{table*}[t]
\centering
\small
\setlength{\tabcolsep}{4pt}
\renewcommand{\arraystretch}{1.35}
\caption{Modeling paradigms for multimodal humor in the large-model era. $x=(x^{v},x^{t})$ denotes the image--text input; $y$ denotes a recognition or interpretation output; $r_{1:K}$ denotes an intermediate rationale chain; $\mathcal{K}_x$ denotes input-specific external knowledge retrieved or selected for $x$; and $c$ denotes a creative control signal.}
\label{tab:paradigm_comparison}
\resizebox{\textwidth}{!}{
\begin{tabular}{>{\raggedright\arraybackslash}p{2.0cm}
                >{\raggedright\arraybackslash}p{2.8cm}
                >{\raggedright\arraybackslash}p{2.6cm}
                >{\raggedright\arraybackslash}p{2.5cm}
                >{\raggedright\arraybackslash}p{2.5cm}
                >{\raggedright\arraybackslash}p{2.2cm}}
\toprule
\rowcolor{headerblue}
\textcolor{white}{\textbf{Paradigm}} & \textcolor{white}{\textbf{Core Technique}} & \textcolor{white}{\textbf{Objectives}} & \textcolor{white}{\textbf{Strength}} & \textcolor{white}{\textbf{Limitation}} & \textcolor{white}{\textbf{Representative}}\\
\midrule
\rowcolor{rowgray}
\multirow{2}{2.0cm}{\textbf{\textcolor{accentblue}{Cross-Modal Alignment}} (\S\ref{sec:alignment})}
& Instruction tuning; contrastive VL pre-training; modular expert routing
& $p_\theta(y \mid x)$: label prediction from a joint visual--textual representation
& Scalable; strong on detection \& classification; transfers across domains
& Mechanism stays implicit; prone to shortcut learning \& hallucinated intent
& SoMeLVLM~\cite{zhang2024somelvlm}, MMoE~\cite{yu2024mmoe}, YesBut-v2~\cite{liang2025yes} \\
\midrule
\multirow{2}{2.0cm}{\textbf{\textcolor{accentblue}{Grounded Reasoning}} (\S\ref{sec:reasoning})}
& CoT prompting; rationale supervision; theory-guided decomposition; external retrieval
& $p_\theta(y,r_{1:K} \mid x,\mathcal{K}_x)$: jointly infer the interpretation and rationale chain, optionally grounded in retrieved knowledge
& Transparent; mechanism-aware explanations; supports cultural, temporal, and social grounding
& Computationally expensive; prompt-sensitive; retrieval noise and reasoning drift can mislead the model
& HumorChain~\cite{zhang2025humorchain}, MemeMind~\cite{gu2025mememind}, MemeX~\cite{sharma2023memex} \\
\midrule
\rowcolor{rowgray}
\multirow{2}{2.0cm}{\textbf{\textcolor{accentblue}{Controllable Generation}} (\S\ref{sec:generation})}
& Prompt-based generation; instruction tuning; control signals; style/content constraints
& $p_\theta(y_{\mathrm{gen}} \mid x,c)$: generate humor-consistent output conditioned on the input and creative control signal
& Enables targeted humor captioning, rewriting, and style-controlled output
& Low controllability; generic outputs; novelty and humor quality hard to evaluate
& MemeCap~\cite{hwang2023memecap}, ViPE~\cite{shahmohammadi2023vipe}, MemeCraft~\cite{wang2024memecraft} \\
\bottomrule
\end{tabular}
}
\vspace{-1em}
\end{table*}

\subsection{Recognition-Oriented Multimodal Alignment}
\label{sec:alignment}
Recognition-oriented systems learn a joint encoder $f_\theta$ that maps visual and textual inputs into a shared semantic space and minimizes a task loss $\mathcal{L} = \mathbb{E}[\ell(g_\phi(f_\theta(x^v, x^t)),\; y)]$.
The critical design choices are (i)~\emph{how} $f_\theta$ is trained---via contrastive pre-training, instruction tuning, or modular routing---and (ii)~\emph{at what granularity} visual features are extracted.
Three strategies span this design space.

\smallskip\noindent\textbf{Creativity-oriented instruction tuning}
adapts a pre-trained MLLM with creativity-specific $(x^v, x^t, \text{instruction}, y)$ supervision.
The principle is that humor, sarcasm, and meme communication have statistical regularities absent from generic VL corpora, so domain-specific fine-tuning consistently outperforms general-purpose checkpoints on recognition benchmarks~\cite{zhang2024somelvlm}.
A contrastive alternative optimizes an InfoNCE loss over image--text embeddings, exposing cross-modal correspondences for multi-task meme classification~\cite{shah2024memeclip}.
Across both strategies, gains are most pronounced on surface-level labels; performance on nuanced intent lags behind, suggesting that alignment alone captures \emph{what co-occurs} but not \emph{why}.

\smallskip\noindent\textbf{Modular and text-centric designs}
decouple perception from reasoning.
Mixture-of-expert routing assigns vision and language tokens to specialized sub-networks, improving robustness when humor depends on only one modality~\cite{yu2024mmoe}.
A more radical approach converts all non-textual modalities into text via captioning, reducing multimodal fusion to unified self-attention~\cite{hasan2023textmi,baluja2025text}.
This text-centric strategy is surprisingly effective for humor understanding---even without architectural changes to the LLM---but incurs information loss on visually dense inputs where spatial layout or fine-grained detail carries the joke.

\smallskip\noindent\textbf{Multi-panel and region-aware alignment}
addresses sequential visual narratives ($x = \{x_1^v, \ldots, x_T^v\}$), where meaning emerges across panels.
The key technical challenge is to capture both intra-panel content and inter-panel relations (entity co-reference, causal flow, setup--punchline structure).
Current approaches range from multi-panel instruction tuning~\cite{liang2025yes}, to panel-selection tasks that test narrative coherence~\cite{vivoli2025comicspap}, to RL-trained region-level encoders that attend to character expressions and speech bubbles within each panel~\cite{chen2025zooming}.
These extensions are necessary because entity continuity and visual salience across panels are prerequisites for downstream interpretation.

\smallskip
Collectively, alignment methods are scalable and strong on classification, but the humor mechanism remains implicit: a model can predict the correct label while failing to recover the violated expectation or social target that drives the humor---a gap confirmed by the recognition-to-interpretation drop in Section~\ref{sec:empirical}.

\subsection{Interpretation-Oriented Reasoning and Grounding}
\begin{figure}
    \centering
    \includegraphics[width=1\linewidth]{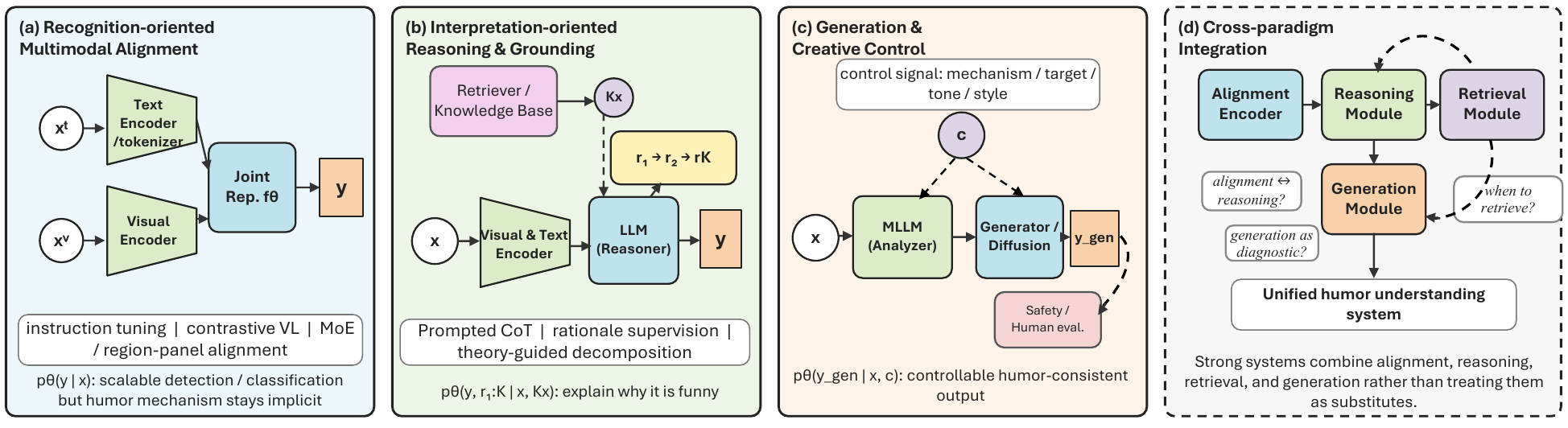}
    \caption{(a) Recognition-oriented multimodal alignment maps visual and textual inputs into a shared representation for scalable humor detection and classification, but leaves the underlying humor mechanism implicit. (b) Interpretation-oriented reasoning and grounding introduces intermediate rationales and optional external knowledge to explain why an artifact is humorous. (c) Generation and creative control conditions humorous output generation on control signals such as mechanism, target, tone, and style, with evaluation and safety constraints. (d) Cross-paradigm integration combines alignment, reasoning, retrieval, and generation into a unified framework, highlighting open questions about alignment–reasoning interaction, retrieval timing, and generation as a diagnostic of understanding.}
    \label{fig:placeholder}
\end{figure}

\label{sec:reasoning}
Once tasks demand explanation, alignment alone is insufficient because it captures \emph{what co-occurs} but not \emph{why it is funny}.
Interpretation-oriented methods address this by conditioning the prediction on a chain of intermediate reasoning steps: $p(y \mid x) \approx \prod_{k=1}^{K} p(r_k \mid r_{<k}, x) \cdot p(y \mid r_{1:K}, x)$,
where each $r_k$ is a natural-language rationale that exposes part of the humor mechanism.
The design space varies along two axes: (i)~\emph{how the reasoning trace is obtained}---prompted at inference, learned from human annotations, or imposed by theory---and (ii)~\emph{where missing knowledge comes from}---the model's own parameters or an external retriever.

\smallskip\noindent\textbf{Prompted vs.\ supervised reasoning.}
The cheapest approach elicits chain-of-thought reasoning at inference time via carefully constructed prompts~\cite{wei2022chain}.
Prompting models to first describe each panel and then articulate the conflict yields clear gains on multi-panel humor, where the contradiction is cross-panel rather than within a single frame~\cite{hu2024cracking}; similar staged prompts decompose conversational jokes into setup, incongruity, and resolution~\cite{chen2024talk}.
However, prompted reasoning is inherently unstable: output quality varies with prompt phrasing, and models can hallucinate plausible-sounding but factually wrong rationales.
Training-time rationale supervision is more robust.
When models are jointly trained to generate the reasoning chain \emph{and} the label---i.e., $p(y, r_{1:K} \mid x)$---both accuracy and interpretability improve substantially over pattern-based baselines, as demonstrated at scale on harmful-meme datasets~\cite{gu2025mememind}.
An alternative is to apply an information bottleneck objective that compresses joke representations to retain only mechanism-relevant features before explanation~\cite{hwang2025bottlehumor}.
The practical trade-off is clear: prompted CoT is zero-cost but fragile; rationale supervision is strong but requires expensive human annotation and risks overfitting to annotator phrasing.

\smallskip\noindent\textbf{Theory-guided decomposition.}
A deeper commitment to structure anchors the reasoning pipeline in established humor theories, imposing fixed stages rather than free-form chains.
The dominant template follows the incongruity-resolution model: (i)~\emph{setup extraction}---identify the expected scenario; (ii)~\emph{conflict detection}---localize the violated expectation; (iii)~\emph{resolution}---explain how the conflict produces humor~\cite{tikhonov2024humor,zhang2025humorchain}.
This staged design generalizes better than single-pass prediction because each stage is independently evaluable, and theory labels (incongruity, superiority, relief) can steer the decomposition.
The same principle of explicit intermediate targets appears in multimodal QA frameworks that chain evidence retrieval with step-by-step reasoning~\cite{agarwal2024mememqa}, in figurative-language benchmarks that frame understanding as natural-language inference over non-literal hypotheses~\cite{saakyan2024v,saakyan2025understanding}, and in probes showing that visual metaphor and symbolic-graphics comprehension collapses without intermediate decomposition~\cite{kundu2025looking,qiu2024can}.
The limitation is that theory-guided pipelines can over-analyze simple humor or introduce reasoning drift when the mechanism does not neatly fit the assumed template.

\smallskip\noindent\textbf{External knowledge retrieval and grounding.}
When the missing information is social, cultural, or temporal rather than perceptual, reasoning over the input alone is insufficient.
Knowledge-grounded systems augment the prediction with a retrieval step: $p(y \mid x, \mathcal{K}_x)$, where $\mathcal{K}_x = \mathcal{R}(x; \mathcal{D})$ is evidence selected from an external corpus $\mathcal{D}$.
Three integration strategies have emerged.
\emph{Dense passage retrieval} encodes the meme as a query and retrieves background documents that explain implicit cultural references~\cite{sharma2023memex}.
\emph{Knowledge-graph and commonsense injection} augments feature representations with structured world knowledge, which is especially valuable for detecting offensiveness that cannot be inferred from surface cues~\cite{garg2025just,kumari2025memedetoxnet}.
\emph{Feature-space adaptation} fuses domain-specific attributes (e.g., brand, persuasion technique) directly into the visual encoder via lightweight adapters, avoiding the latency of an explicit retriever~\cite{jia2023kafa}.
A related technique retrieves similar labeled examples at inference time for in-context learning, enabling domain-shift robustness without retraining~\cite{tang2024leveraging};
reasoning-knowledge distillation from large LLMs into smaller students offers another route to grounding~\cite{lin2023beneath}.
Retrieval is critical for satire, time-sensitive memes, and culturally loaded references, but it introduces noise: irrelevant evidence can mislead the model and obscure the actual creative mechanism.

\subsection{Generation and Creative Control}
\label{sec:generation}
Generation can be framed as conditional decoding $p(y_{\text{gen}} \mid x, c)$, where $x$ is the source artifact and $c$ encodes a creative control signal (intended mechanism, target concept, tone).
The central difficulty is that effective $c$ presupposes interpretive competence: generating a funny caption for an image requires understanding \emph{what is funny about} that image.
Current work separates along three design principles.

\smallskip\noindent\textbf{Mechanism-explicit captioning}
adopts a two-stage \emph{analyze-then-generate} pipeline.
The analysis stage identifies the incongruity, salient objects, or violated expectation; the generation stage decodes a caption conditioned on that analysis.
This architecture---instantiated through incongruity-resolution CoT prompting~\cite{tanaka-etal-2024-content}, grounded humor captioning benchmarks~\cite{li2023oxfordtvg}, and cascaded describe-explain-caption pipelines~\cite{hwang2023memecap}---consistently outperforms direct prompting on image-specificity.
A practical finding is that decoding-time interventions such as logit bias and negative sampling further suppress generic outputs, confirming that the bottleneck is not fluency but groundedness.
The cascaded design has its own cost: errors in early stages propagate, making the overall quality sensitive to the weakest link.

\smallskip\noindent\textbf{Creative artifact synthesis}
extends generation beyond text to novel images or multimodal artifacts, typically by separating semantic planning (LLM) from visual rendering (diffusion model).
Visual metaphor generation pipelines first extract source and target domains, then compose a concrete scene description, and finally render it---allowing each module to be evaluated and improved independently~\cite{chakrabarty2023spy,shahmohammadi2023vipe}.
Meme generation adds a stance or safety dimension: unconstrained generation produces more novel outputs but also higher rates of harmful content~\cite{wang2024memecraft}.
A distinct thread explores ``creative leaps''---humor requiring associative jumps beyond logical entailment.
The emerging finding is that moderate-distance leaps between setup and resolution are rated funniest by human judges; too close is predictable, too far is nonsensical~\cite{zhong2024let,wang2024innovative}.

\smallskip\noindent\textbf{Human-AI co-creation}
addresses a key limitation of fully autonomous systems: the absence of audience feedback during creative production.
Interactive tools that suggest incongruity-based options for human refinement show that even novice users produce higher-quality humor when the system handles mechanism generation while the human curates tone and relevance~\cite{kariyawasam2024appropriate}.
Extending generation to temporal media such as short-video commentary further raises the bar, requiring both cross-modal alignment and temporal reasoning~\cite{ouyang2025laugh}.

\smallskip
Generation remains bottlenecked by controllability and evaluation.
Models often regress toward generic captions or reuse familiar templates.
Current metrics---both reference-based (BLEU, BERTScore) and human preference---cannot reliably distinguish genuine novelty from paraphrasing.
Generation is therefore best viewed as a stress test of interpretive competence: without understanding the mechanism, faithful creative output is unlikely.

\subsection{Cross-Paradigm Analysis}
The three paradigms are complementary: alignment provides scalable perception, reasoning adds interpretive depth, and retrieval supplies missing context.
The strongest recent systems already combine elements---e.g., theory-guided reasoning over alignment-tuned features~\cite{zhang2025humorchain}, CoT supervision with commonsense grounding~\cite{gu2025mememind}, or reasoning decomposition feeding into generation~\cite{tanaka-etal-2024-content}---but integration remains ad hoc rather than principled.

Three open questions emerge from this landscape.
First, \emph{alignment--reasoning interaction}: does finer-grained visual alignment (region-level, panel-level) reduce the burden on downstream reasoning, or does it introduce redundant detail that increases reasoning drift?
Second, \emph{retrieval timing}: should evidence be injected before reasoning (to inform the trace) or after an initial reasoning pass (to fill identified gaps)?
Current systems use fixed timing, but adaptive strategies could improve efficiency.
Third, \emph{generation as a diagnostic for understanding}: the field treats generation as a downstream application, yet generation quality could serve as a stronger probe of interpretive competence than MCQ accuracy---if appropriate evaluation protocols existed.
Addressing these questions will require joint benchmarks that evaluate alignment, reasoning, and generation on the same artifacts, a direction we revisit in Section~\ref{sec:empirical}.

\section{Datasets and Benchmarks}
\label{sec:datasets}

\begin{table*}[t]
\centering
\small
\setlength{\tabcolsep}{5pt}
\renewcommand{\arraystretch}{1.3}
\caption{Capability-aligned summary of benchmark resources. Detailed dataset inventories appear in Appendix Tables~\ref{tab:reco-datasets1} and \ref{tab:datasets3}.}
\label{tab:dataset_by_capability_level}
\begin{tabularx}{\linewidth}{
  >{\raggedright\arraybackslash}p{2.0cm}
  Z
  Y
  Y
  T}
\toprule
\rowcolor{headerblue}
\textcolor{white}{\textbf{Level}} & \textcolor{white}{\textbf{Representative Resources}} & \textcolor{white}{\textbf{Typical Supervision / Target}} & \textcolor{white}{\textbf{Typical Evaluation}} & \textcolor{white}{\textbf{Recurring Limitation}} \\
\midrule
\rowcolor{rowgray}
\textbf{\textcolor{accentblue}{Recognition}}
& RedEval~\cite{tang2024leveraging}, HumorDB~\cite{jain2025humordb}, COMICORDA~\cite{martinek2024comicorda}, Inside Jokes~\cite{shahaf2015inside}
& Binary labels, role labels, dialogue acts, or intensity scores
& Accuracy, F1, AUROC, correlation
& Shortcut learning remains easy; labels reveal little about interpretive depth. \\
\midrule
\textbf{\textcolor{accentblue}{Interpretation \& reasoning}}
& Do Androids Laugh at Electric Sheep?~\cite{hessel2023androids}, V-FLUTE~\cite{saakyan2025understanding}, PixelHumor~\cite{ryan2025humor}, YesBut~\cite{hu2024cracking}
& Explanations, rationales, answer selection, cross-panel reasoning targets
& MCQ accuracy, BERTScore, rubric-based human or LLM-assisted judgment
& Limited cultural coverage and weak evidence annotation make faithful evaluation difficult. \\
\midrule
\rowcolor{rowgray}
\textbf{\textcolor{accentblue}{Generation}}
& MEMECAP~\cite{hwang2023memecap}, OxfordTVG-HIC~\cite{li2023oxfordtvg}, Oogiri-GO~\cite{zhong2024let}, XMeCap~\cite{chen2024xmecap}
& Captions, punchlines, continuations, or humorous rewrites
& Reference overlap, LLM-as-judge, human preference
& Novelty, faithfulness, and safety are hard to score automatically; dedicated benchmarks remain sparse. \\
\bottomrule
\end{tabularx}
\end{table*}

In this section, we align benchmarks with the same capability hierarchy used for tasks and models. This makes it easier to distinguish resources that only test label prediction from those that require explanation, cross-panel inference, or image-grounded generation. Detailed inventories are provided in Appendix Tables~\ref{tab:reco-datasets1} and  \ref{tab:datasets3}.

\smallskip
\noindent\textbf{Recognition Resources.} 
Recognition benchmarks remain the most abundant and the most standardized. In the image-based setting, they cover humorous cartoons, sarcastic image--text pairs, and multi-panel comics through labels for humor presence, target type, dialogue act, or intensity~\cite{shahaf2015inside,tang2024leveraging,jain2025humordb,martinek2024comicorda}. These resources are useful for training perception and alignment modules, but they remain classification-oriented and therefore provide only indirect evidence about whether a model understands the underlying joke, target, or narrative conflict.


\smallskip
\noindent\textbf{Interpretation and Reasoning Resources.} More recent benchmarks explicitly probe interpretive depth by pairing image-based inputs with explanations, rationales, multiple-choice reasoning questions, or cross-panel inference targets. Datasets such as \textit{Do Androids Laugh at Electric Sheep?}, V-FLUTE, PixelHumor, and the YesBut series move beyond surface labels and ask whether models can identify why a caption is funny, what contradiction drives a panel sequence, or how an implicit target should be resolved~\cite{hessel2023androids,saakyan2025understanding,hu2024cracking,ryan2025humor}. These resources are far more diagnostic than recognition datasets, but they remain smaller, culturally narrower, and more expensive to annotate.

\smallskip
\noindent\textbf{Generation Resources.} Dedicated generation benchmarks are fewer and often repurpose understanding data as supervised targets. MEMECAP and OxfordTVG-HIC center humorous caption generation from static images, while Oogiri-GO and XMeCap extend the space toward joke completion, meme rewriting, or continuation over structured inputs~\cite{hwang2023memecap,li2023oxfordtvg,zhong2024let,chen2024xmecap}. Their main value is diagnostic: they reveal whether a model can transform an interpretation of the source image into a controlled, novel output. At present, however, generation benchmarks remain sparse, and their evaluation protocols are still less mature than those used for recognition or explanation.

\section{Cross-Benchmark Empirical Analysis}
\label{sec:empirical}
\begin{table*}[t]
\centering
\scriptsize
\setlength{\tabcolsep}{3pt}
\caption{
Results on humor understanding benchmarks. All numbers are accuracy (\%). Model results are obtained by evaluating a broader and more recent set of MLLMs using the benchmark-specific prompts and task definitions reported in the original papers. Human performance figures, where available, are taken from the corresponding benchmark papers. Task abbreviations: for \textbf{YesBut-v2}~\cite{liang2025yes}, \textit{Moral} and \textit{Title} denote choosing the correct moral or title from multiple choices; 
\textbf{NYCC}~\cite{hessel2023androids} denotes selecting the most suitable \textit{Caption} for a New Yorker cartoon from options A--E; 
\textbf{MemeQA}~\cite{nguyen2025memeqa} denotes multiple-choice \textit{Fill}-in-the-blank question answering on memes; 
\textbf{ExHVV}~\cite{sharma2023you} denotes \textit{Role} selection for entities from \{hero, villain, victim\} in memes; 
\textbf{DarkHumor}~\cite{kasu2025d} denotes binary \textit{Detect}ion of dark humor (Yes/No); 
\textbf{HumorDB}~\cite{jain2025humordb} denotes binary \textit{Detect}ion of humor (Yes/No). 
For \textbf{MangaUB}~\cite{ikuta2025mangaub}, \textit{RecBg}, \textit{CharCnt}, \textit{PanelLoc}, \textit{NextInf}, and \textit{Onom} denote \textit{recognition\_background}, \textit{character\_count}, \textit{panel\_localization}, \textit{next\_panel\_inference}, and \textit{onomatopoeia\_scene}, respectively.
}
\label{tab:humor_benchmarks}
\resizebox{\textwidth}{!}{
\begin{tabular}{lcccccccccccc}
\toprule
\multirow{3}{*}{\textbf{Models}} 
& \multicolumn{8}{c}{\textbf{Recognition}} 
& \multicolumn{4}{c}{\textbf{Interpretation \& Reasoning}} \\
\cmidrule(lr){2-9}
\cmidrule(lr){10-13}
& \textbf{ExHVV} 
& \textbf{DarkHumor} 
& \textbf{HumorDB} 
& \multicolumn{5}{c}{\textbf{MangaUB}} 
& \multicolumn{2}{c}{\textbf{YesBut-v2}} 
& \textbf{NYCC} 
& \textbf{MemeQA} \\
\cdashline{2-2}
\cdashline{3-3}
\cdashline{4-4}
\cdashline{5-9}
\cdashline{10-11}
\cdashline{12-12}
\cdashline{13-13}
& \textit{Class.} 
& \textit{Detect} 
& \textit{Detect} 
& \textit{RecBg} 
& \textit{CharCnt} 
& \textit{PanelLoc} 
& \textit{NextInf} 
& \textit{Onom} 
& \textit{Moral} 
& \textit{Title} 
& \textit{Caption} 
& \textit{Fill} \\
\midrule
Qwen2.5-VL-7B         & 71.65 & 47.37 & 65.90 & 91.69 & 85.20 & 79.19 & 35.01 & 87.21 & 67.33 & 76.94 & 47.34 & 39.61 \\
LLaVA-OneVision-7B    & 73.48 & 48.06 & 64.06 & 97.57 & 93.79 & 67.59 & 33.37 & 83.66 & 67.64 & 70.80 & 58.14 & 40.96 \\
Qwen3-VL-8B           & 76.24 & 49.43 & 71.02 & 95.10 & 92.02 & 83.64 & 48.55 & 89.03 & 74.69 & 80.51 & 50.94 & 49.84 \\
Qwen3-VL-8B-Thinking  & 78.64 & \textbf{66.43} & 70.59 & 94.35 & 90.07 & 82.26 & 32.76 & 91.34 & 70.67 & 78.82 & 46.49 & 37.66 \\
InternVL3.5-8B        & 77.64 & 49.77 & 72.58 & 94.97 & 90.87 & 78.58 & 46.68 & 88.45 & 69.30 & 76.10 & 48.86 & 49.50 \\
\midrule
InternVL3.5-30B-A3B   & 79.44 & 54.76 & 70.45 & 96.59 & 89.98 & 82.06 & 48.22 & 90.10 & 75.46 & --    & 51.89 & 49.40 \\
Qwen3.5-35B-A3B       & \textbf{80.03} & 66.37 & 68.60 & 92.90 & 91.22 & 87.27 & 33.46 & 84.57 & 76.06 & 78.40 & 60.25 & \textbf{60.41} \\
\midrule
Gemma-4-31B-it        & 69.35 & 57.68 & 51.10 & 39.20 & 12.85 & 10.01 & 23.51 & 63.20 & 52.71 & 35.65 & 22.31 & 40.16 \\
Qwen3.5-27B           & 77.24 & 65.84 & \textbf{74.71} & 94.04 & \textbf{95.21} & 90.95 & 54.08 & 83.25 & \textbf{84.72} & \textbf{83.28} & 61.79 & 54.52 \\
InternVL3.5-38B       & 70.06 & 61.47 & 51.98 & 69.16 & 12.94 & 41.72 & 45.43 & 69.64 & 64.58 & 69.65 & 34.81 & 51.39 \\
\midrule
GPT-4o                & 74.80 & 61.50 & 59.95 & \textbf{97.60} & 94.50 & \textbf{99.21} & \textbf{64.30} & \textbf{95.80} & 80.38 & 80.62 & \textbf{82.30} & 59.60 \\
\rowcolor{blue!8}
Human                 & 81.00 & --    & 85.00 & \multicolumn{5}{c}{--} & 91.30 & 97.50 & 94.00 & 81.90 \\
\bottomrule
\end{tabular}
}
\end{table*}

To move beyond qualitative synthesis, we extend the original evaluation
protocols of the surveyed benchmarks to a broader and more recent set of
MLLMs, while retaining the benchmark-specific prompts reported in the
corresponding papers. We organize the resulting scores according to the
capability hierarchy introduced in Section~3. This cross-benchmark empirical
analysis serves two purposes: (1) it provides an updated snapshot of current
MLLM performance across different humor-understanding capabilities, and
(2) it reveals recurring patterns that are difficult to observe from individual
benchmarks alone, particularly the contrast between recognition-oriented tasks
and interpretation- or reasoning-oriented tasks. Human performance figures,
where available, are taken from the original benchmark papers.

\paragraph{Evaluation protocol.}
For each benchmark, we follow the prompt format and task definition reported
in the original paper. We use the official evaluation split whenever available
and evaluate a broader set of recent MLLMs under the same benchmark-specific
protocol. Because prompts differ across benchmarks by design, the results
support within-benchmark model comparison and descriptive cross-benchmark
analysis, rather than a strictly controlled comparison of task difficulty.

\subsection{Cross-Level Performance Landscape}

Table~\ref{tab:humor_benchmarks} summarizes MLLM results across seven
humor-understanding benchmarks and twelve task settings. Following our
capability hierarchy, we group ExHVV, DarkHumor, HumorDB, and MangaUB under
\textit{recognition}, since these tasks primarily require models to identify
roles, detect humor categories, or recognize visual and structural elements.
We group YesBut-v2, NYCC, and MemeQA under
\textit{interpretation \& reasoning}, since these tasks require models to infer
morals, select appropriate titles or captions, and fill in missing semantic
content. 

Several patterns emerge from this cross-benchmark view.

\paragraph{Recognition-oriented tasks generally yield higher reported
accuracies, but remain far from uniformly solved.}
Models generally perform strongly on visually grounded recognition tasks,
especially the MangaUB subtasks. GPT-4o achieves 97.60\% on background
recognition, 99.21\% on panel localization, 64.30\% on next-panel inference,
and 95.80\% on onomatopoeia-scene recognition, obtaining the best score on
four of the five MangaUB subtasks. Open-source models are also competitive on
several recognition tasks: Qwen3.5-27B reaches 95.21\% on character counting
and 74.71\% on HumorDB, while Qwen3.5-35B-A3B achieves the best ExHVV
role-classification accuracy at 80.03\%. However, recognition is not uniformly
easy. DarkHumor remains difficult for most models, with the best score reaching
only 66.43\% with Qwen3-VL-8B-Thinking. This suggests that recognition remains
challenging when it depends on implicit social norms, taboo framing, or
pragmatic cues.

\paragraph{Interpretation and reasoning remain the central bottleneck.}
Compared with recognition-oriented tasks, interpretation-oriented benchmarks
show larger and more consistent gaps from human performance. On YesBut-v2,
the best model score is 84.72\% for moral selection and 83.28\% for title
selection, both achieved by Qwen3.5-27B, while human performance reaches
91.30\% and 97.50\%, respectively. On NYCC caption selection, GPT-4o
substantially outperforms all other models with 82.30\%, but still trails the
human score of 94.00\%. MemeQA shows a similar gap: the best model,
Qwen3.5-35B-A3B, reaches 60.41\%, compared with 81.90\% for humans. These
results indicate that current MLLMs can often recognize salient entities or
visual structures, but still struggle to infer the intended humorous mechanism,
implicit punchline, or culturally appropriate interpretation.

\paragraph{Model ranking varies substantially across humor capabilities.}
No single model dominates all task types. GPT-4o is strongest on NYCC and
most MangaUB subtasks, suggesting strong visual recognition and
caption-matching ability. Qwen3.5-27B performs best on both YesBut-v2
subtasks and HumorDB, indicating stronger performance on moral and title
inference as well as general humor detection. Qwen3.5-35B-A3B achieves the
highest ExHVV and MemeQA scores, while Qwen3-VL-8B-Thinking performs best
on DarkHumor. This fragmented ranking suggests that humor understanding is
not a single monolithic ability. Different benchmarks emphasize different
combinations of visual recognition, social knowledge, narrative inference,
and pragmatic reasoning.

\paragraph{Reasoning-oriented variants do not consistently improve humor
understanding.}
The comparison between Qwen3-VL-8B and Qwen3-VL-8B-Thinking is
particularly revealing. The Thinking variant substantially improves
DarkHumor detection from 49.43\% to 66.43\% and improves ExHVV from
76.24\% to 78.64\%, suggesting that deliberative reasoning can help when the
task requires social or normative judgment. However, it decreases performance
on several interpretation tasks, including YesBut-v2 Moral, NYCC, and MemeQA,
and also drops sharply on MangaUB next-panel inference. This mixed pattern
indicates that explicit reasoning does not automatically translate into better
humor understanding. In some settings, extended reasoning may help identify
implicit intent, while in others it may divert the model from direct
visual--semantic matching or introduce unnecessary intermediate steps.

\paragraph{Human performance remains substantially higher on
interpretation-heavy tasks.}
Where human results are available, the largest gaps appear in tasks requiring
semantic or pragmatic interpretation. Humans outperform the best model by
6.58 points on YesBut-v2 Moral, 14.22 points on YesBut-v2 Title, 11.70 points
on NYCC, and 21.49 points on MemeQA. The gap is also large on HumorDB, where
the best model reaches 74.71\% compared with 85.00\% for humans. These gaps
reinforce the main finding of this section: current MLLMs have made
considerable progress on visual and categorical recognition, but still fall
short in recovering the intended meaning, communicative function, and
incongruity structure that make visual humor understandable to humans.

\section{Challenges and Future Directions}
\label{sec:challenges}

Despite rapid progress in MLLMs on literal scene perception, a substantial gap remains between recognizing \emph{what is depicted} and interpreting \emph{what is meant} in multimodal visual humor. Closing this gap requires moving beyond physical description toward socio-cultural, rhetorical, and value-laden interpretation. We synthesize this gap into  interconnected challenges and outline concrete directions for the community.

 








\subsection{The Evaluation Crisis}

Most benchmarks reduce humor understanding to multiple-choice questions (MCQs) or binary classification~\cite{hessel-etal-2023-androids,hu2024cracking,yang-etal-2024-large}. This is scalable but poorly matched to the phenomenon: discriminative accuracy is inflated by shortcut learning, and creative interpretation is inherently open-ended.

\smallskip\noindent\textbf{Future directions.}
A promising direction is \emph{rubric-based generative evaluation}, where models produce free-form interpretations that are assessed along disentangled dimensions such as incongruity detection, target identification, and cultural grounding, rather than a single aggregate label~\cite{gunjal2025rubrics,liang2025yes}. This can be coupled with LLM-as-judge frameworks to operationalize humor-specific rubrics, rewarding plausible novel interpretations while penalizing unsupported or hallucinated reasoning~\cite{liu2025inference,liu2023g}. Such a paradigm better aligns evaluation with the inherently open-ended and interpretive nature of humor understanding.

\subsection{Social and Cultural Grounding}

Creative artifacts depend on shared background knowledge, implicit norms, and audience beliefs that current MLLMs reason about poorly~\cite{jiang2025can,hu2023language,chiu2025culturalbench}. First, current large models still lack theory-of-mind–like reasoning to infer whose perspective is expressed, what belief is being challenged, how an audience is meant to react, and often default to flat literal description instead~\cite{chen-etal-2025-theory}. In addition, parametric knowledge has a fixed cutoff, so memes tied to breaking news or ephemeral trends become opaque~\cite{kasai2024realtime}, while training data remains predominantly English and Western, producing systematic blind spots on non-Western visual symbols and humor conventions~\cite{yu2025seeing,park2025evaluating}. 

Early efforts such as CHumor for Chinese humor~\cite{he2024chumor}, Chinese multimodal sarcasm~\cite{gao2025multimodal}, and StandUp4AI for multilingual stand-up~\cite{barriere2025standup4ai} broaden coverage, but still capture only a small slice of global humor traditions. Retrieval-augmented generation (RAG)~\cite{lewis2020retrieval} offers a promising path: retrieval-guided learning improves hateful-meme detection~\cite{mei2024improving} and contextualized meme explanation~\cite{sharma2023memex}, yet humor demands retrieval of not only topical knowledge but also the social norms and cultural contexts that make jokes intelligible.

\smallskip\noindent\textbf{Future directions.}
(i)~\emph{Multicultural, per-culture benchmarks} with annotations of the background knowledge each item requires, rather than a single aggregate accuracy;
(ii)~\emph{humor-aware RAG}: retrieval pipelines that surface community norms, trending discourse, and event timelines alongside factual context, paired with periodic knowledge refresh to mitigate temporal decay;
(iii)~\emph{explicit intent and stance modeling} (critique, mockery, self-deprecation) as a structured, trainable proxy for theory-of-mind reasoning.

\subsection{Narrative Reasoning in Sequential Humor}

Our cross-benchmark analysis (Section~\ref{sec:empirical}) shows that sequential, multi-panel humor exposes the widest model–human gap on panel-sequencing and temporal reordering~\cite{ryan2025humor,wang-etal-2025-beyond-single}. Unlike single-image humor, where the incongruity sits within one frame, multi-panel humor requires maintaining entity identity across panels, building an expectation from setup panels, and localizing the exact point where the narrative violates it. Current architectures largely process panels in isolation, missing these temporal and causal dependencies.

\smallskip\noindent\textbf{Future directions.}
(i)~\emph{Panel-aware architectures} that explicitly encode panel order and cross-frame entity co-reference, borrowing temporal-attention and state-tracking inductive biases from video understanding;
(ii)~\emph{setup–punchline decomposition} as an explicit training objective rather than holistic pattern matching;
(iii)~\emph{diagnostic benchmarks} isolating individual narrative skills (next-panel prediction, swapped-panel detection) for targeted evaluation beyond aggregate accuracy.

\subsection{Toward Unified, Safe Systems}

Recognition, interpretation, and generation are currently treated as independent tasks with separate pipelines, despite being deeply intertwined. A system that can explain why a meme is funny should, in principle, be better at generating one. This fragmentation is compounded by a safety problem specific to humor: the same non-literal mechanisms that make humor effective (incongruity, exaggeration, irony) are also what makes harmful content hard to detect. The Hateful Memes Challenge showed that neither unimodal classifier can reliably catch cross-modally-constructed hate~\cite{kiela2020hateful}, and follow-up work shows detecting \emph{who} is targeted is as important as detecting \emph{that} harm occurs~\cite{sharma2022disarm,pramanick2021momenta}. As generation models improve, this becomes dual-use: the same system that writes witty captions can be steered toward harassment or stereotype-reinforcing content~\cite{weidinger2022taxonomy,wang2024memecraft}, and current VLMs remain vulnerable to adversarial meme-based attacks~\cite{lee2025vision}. Overly conservative filters, in turn, risk censoring legitimate satire and self-deprecating humor — an unresolved safety–creativity trade-off~\cite{jha2024memeguard}. Separately, creative datasets are frequently scraped without creator consent, raising unresolved data-provenance and ``right to style'' concerns as models move toward style-imitative generation.

\smallskip\noindent\textbf{Future directions.}
(i)~\emph{Joint training across the capability hierarchy} (recognition, interpretation, generation) to test whether gains at one level transfer to the others, and \emph{self-consistency checks} (e.g., a model that rates a meme funny but explains it blandly) as a diagnostic for genuine understanding;
(ii)~\emph{context-sensitive, rhetoric-aware moderation}: safety classifiers conditioned on whether an artifact critiques versus promotes a harmful stance, paired with humor-specific red-teaming rather than keyword filtering;
(iii)~\emph{provenance-aware training}: datasets and models that record content origin, licensing, and consent, auditable back to their training sources.

\section{Conclusion}
Multimodal visual humor challenges AI systems as its meaning extends beyond surface perception. Achieving robust understanding and generation therefore requires moving toward socially and rhetorically grounded reasoning. This survey introduces a capability-centric task hierarchy that clarifies how creative understanding progresses from recognition to interpretation and generation. We further outline future directions for meaningful, reliable, and responsible engagement with human-created media.


\section*{Limitations}

Despite growing interest in multimodal humor, this survey has several
limitations. First, most existing research and datasets focus on Western,
internet-centric visual humor forms (e.g., memes, cartoons), leaving many
cultural traditions and non-mainstream media underexplored. Second, as
multimodal models evolve rapidly, some observations may not fully generalize
to future architectures or training paradigms.


\section*{Ethical considerations}

Understanding creative expression poses distinct ethical challenges. Creative
media may potentially convey harmful or sensitive content implicitly through
humor, irony, or symbolism, increasing the risk of misinterpretation, bias
amplification, or over-censorship. Dataset bias and cultural imbalance further
threaten fairness and robustness, particularly for marginalized communities.
In addition, many creative datasets raise unresolved copyright and ownership
concerns, especially as models increasingly transition from understanding to
generation and style imitation. Addressing these issues requires context-aware
evaluation, transparent dataset practices, and greater emphasis on
interpretability and human oversight when deploying such systems.


\bibliographystyle{bibstyle}
\bibliography{custom}


\clearpage
\appendix
\section{Evaluation Details}
\label{app:evaluation_details}

We evaluate all models in a zero-shot setting. For each benchmark, we retain
the task definition, prompt format, answer format, evaluation split, and scoring
procedure reported in the corresponding original paper. Except for GPT-4o,
all evaluated models use publicly available checkpoints hosted on Hugging
Face. For GPT-4o, we use the fixed API version
\texttt{gpt-4o-2024-05-13}. All experiments were completed by May 29, 2026.

Each model is queried once per question. For open-source models, decoding is
performed with \texttt{do\_sample=true}; all remaining benchmark-specific
generation and evaluation settings follow the corresponding original papers.
When an official evaluation split is available, we use it directly; otherwise,
we follow the split or sampling procedure described in the benchmark paper.
Because prompts and evaluation protocols differ across benchmarks by design,
the reported results are intended primarily for within-benchmark model
comparison and descriptive cross-benchmark analysis, rather than for a strictly
controlled comparison of task difficulty. Since each question is evaluated
with a single sampled generation, small score differences may be affected by
decoding stochasticity and should not be interpreted as statistically
significant.

\section{Overview of Datasets and Benchmarks}
Here we provide a comprehensive tabular overview of datasets and benchmarks studied in this survey, organized according to the capability hierarchy introduced in the main paper.

\begin{table*}[t]
\centering
\small
\setlength{\tabcolsep}{4pt}
\renewcommand{\arraystretch}{1.25}
\caption{Overview of multimodal visual humor datasets focused on \textbf{Recognition} tasks. Recognition datasets typically use binary labels (e.g., humorous vs.\ non-humorous), element-level labels (e.g., punchlines or rhetorical roles), and intensity scores (e.g., degree of funniness or offensiveness). In Data Forms, StaVT denotes Static Visual–Textual Artifacts; SeqVN denotes Sequential Visual Narratives. In Mechanism, Multi denotes datasets that involve multiple mechanisms defined in Sec.~\ref{taxonomy}. The Availability column provides links to publicly accessible datasets, and "N/A" indicates unpublished datasets.}
\begin{tabularx}{\linewidth}{
  >{\raggedright\arraybackslash}X
  >{\raggedright\arraybackslash}p{1.2cm}
  >{\raggedright\arraybackslash}p{2.0cm}
  >{\raggedright\arraybackslash}p{1.3cm}
  >{\raggedleft\arraybackslash}p{0.9cm}
  >{\centering\arraybackslash}p{0.8cm}}
\toprule
\rowcolor{headerblue}
\textcolor{white}{\textbf{Dataset}} &
\textcolor{white}{\textbf{Venue}} &
\textcolor{white}{\textbf{Data Forms}} &
\textcolor{white}{\textbf{Mechanism}} &
\textcolor{white}{\textbf{Size}} &
\textcolor{white}{\textbf{Avail.}} \\
\midrule

\rowcolor{groupA}
\multicolumn{6}{l}{\textit{\textcolor{headerblue}{\textbf{StaVT: Meme}}}} \\
\rowcolor{groupA}
\multicolumn{6}{l}{\textit{\textcolor{headerblue}{\textbf{Goal: Humor \& Entertainment, Satire \& Social Critique}}}} \\
\rowcolor{groupA} \textbf{D-HUMOR}~\cite{kasu2025d} & ICDM'25 & StaVT: Meme & Multi & 4{,}379 & \href{https://github.com/Sai-Kartheek-Reddy/D-Humor-Dark-Humor-Understanding-via-Multimodal-Open-ended-Reasoning}{Link} \\
\rowcolor{groupA} \textbf{MemeMind}~\cite{gu2025mememind} & Arxiv'25 & StaVT: Meme & Multi & 43{,}223 & \textcolor{naGray}{\textit{N/A}} \\
\rowcolor{groupA} \textbf{TOXICN MM}~\cite{lu2024towards} & NeurIPS'24 & StaVT: Meme & Multi & 12K & \href{https://github.com/DUT-lujunyu/ToxiCN_MM}{Link} \\
\rowcolor{groupA} \textbf{PrideMM}~\cite{shah2024memeclip} & EMNLP'24 & StaVT: Meme & Multi & 5{,}063 & \href{https://drive.google.com/file/d/17WozXiXfq44Z6kkWsPPDHRzqIH2daUaQ/view}{Link} \\
\rowcolor{groupA} \textbf{BHM}~\cite{hossain2024deciphering} & ACL'24 & StaVT: Meme & Multi & 7{,}148 & \textcolor{naGray}{\textit{N/A}} \\
\rowcolor{groupA} \textbf{Ext-Harm-P}~\cite{sharma2022disarm} & NAACL'23 & StaVT: Meme & Multi & 4{,}446 & \href{https://github.com/LCS2-IIITD/DISARM}{Link} \\
\rowcolor{groupA} \textbf{HVVMemes}~\cite{sharma2023characterizing} & EACL'23 & StaVT: Meme & Multi & 6{,}933 & \href{https://codalab.lisn.upsaclay.fr/competitions/906}{Link} \\
\rowcolor{groupA} \textbf{RESTORE}~\cite{yadav2023towards} & ACL'23 & StaVT: Meme & Multi & 4{,}664 & \href{https://docs.google.com/forms/d/e/1FAIpQLSeK8h2TvDt-XzZzvy0vWvjauDAX8eUXJ6OUuoAfd4ZqWQPqLA/viewform}{Link} \\
\rowcolor{groupA} \textbf{Dank or not?}~\cite{Barnes2020DankON} & App.\ Net.\ Sci.'21 & StaVT: Meme & Multi & 70K & \textcolor{naGray}{\textit{N/A}} \\
\rowcolor{groupA} \textbf{The Hateful Memes Challenge Set}~\cite{kiela2020hateful} & NeurIPS'20 & StaVT: Meme & Multi & 10K & \href{https://www.drivendata.org/competitions/64/hateful-memes/}{Link} \\

\midrule
\rowcolor{groupB}
\multicolumn{6}{l}{\textit{\textcolor{headerblue}{\textbf{StaVT: Sarcastic Image}}}} \\
\rowcolor{groupB}
\multicolumn{6}{l}{\textit{\textcolor{headerblue}{\textbf{Goal: Satire \& Social Critique}}}} \\
\textbf{RedEval}~\cite{tang2024leveraging} & NAACL'24 & StaVT: Sarcastic Image & Sarcasm & 1{,}004 & \href{https://github.com/TangBinghao/naacl2024}{Link} \\
\textbf{SPMSD}~\cite{guo2025multi} & COLING'24 & StaVT: Sarcastic Image & Sarcasm & 1K & \textcolor{naGray}{\textit{N/A}} \\
\textbf{MSTI dataset}~\cite{wang2022multimodal} & ACL'22 & StaVT: Sarcastic Image & Sarcasm & 5{,}015 & \href{https://github.com/wjq-learning/MSTI}{Link} \\
\textbf{Multi-Modal Sarcasm Detection in Twitter}~\cite{cai2019multi} & ACL'19 & StaVT: Sarcastic Image & Sarcasm & 24{,}635 & \textcolor{naGray}{\textit{N/A}} \\
\textbf{Sarcasm in Multimodal Social Platforms}~\cite{schifanella2016detecting} & ACM MM'16 & StaVT: Sarcastic Image & Sarcasm & 10K & \textcolor{naGray}{\textit{N/A}} \\

\midrule
\rowcolor{groupC}
\multicolumn{6}{l}{\textit{\textcolor{headerblue}{\textbf{StaVT: Humorous Image / Cartoon}}}} \\
\rowcolor{groupC}
\multicolumn{6}{l}{\textit{\textcolor{headerblue}{\textbf{Goal: Humor \& Entertainment}}}} \\
\rowcolor{groupC} \textbf{HumorDB}~\cite{jain2025humordb} & ICCV'25 & StaVT: Humorous Image & Multi & 3{,}542 & \href{https://github.com/kreimanlab/HumorDB}{Link} \\
\rowcolor{groupC} \textbf{Inside Jokes}~\cite{shahaf2015inside} & KDD'15 & StaVT: Cartoon & Multi & 76{,}928 & \textcolor{naGray}{\textit{N/A}} \\

\midrule
\rowcolor{groupD}
\multicolumn{6}{l}{\textit{\textcolor{headerblue}{\textbf{SeqVN: Comic Strip}}}} \\
\rowcolor{groupD}
\multicolumn{6}{l}{\textit{\textcolor{headerblue}{\textbf{Goal: Humor \& Entertainment}}}} \\
\textbf{COMICORDA}~\cite{martinek2024comicorda} & COLING'24 & SeqVN: Comic Strip & Narrative & 1{,}438 & \textcolor{naGray}{\textit{N/A}} \\
\textbf{AVH \& FOR}~\cite{chandrasekaran2016we} & CVPR'16 & SeqVN: Comic Strip & Narrative & 7{,}150 & \href{https://github.com/GT-Vision-Lab/abstract_scenes_v002}{Link} \\

\bottomrule
\end{tabularx}
\label{tab:reco-datasets1}
\end{table*}

\begin{table*}[t]
\centering
\small
\setlength{\tabcolsep}{4pt}
\renewcommand{\arraystretch}{1.25}
\caption{Overview of multimodal visual humor datasets focused on \textbf{Understanding} and \textbf{Generation}. These datasets typically pair multimodal inputs with explanations or rationales—often augmented with contextual or external knowledge—to justify intended meaning (e.g., humor, irony, or satire), alongside target outputs that support coherent and controllable generation (e.g., memes or comics with captions/rationales). StaVT denotes Static Visual–Textual Artifacts; SeqVN denotes Sequential Visual Narratives. In Mechanism, Multi denotes datasets that involve multiple mechanisms defined in Sec.~\ref{taxonomy}. The Availability column provides links to publicly accessible datasets, and "N/A" indicates unpublished datasets.}
\vspace{-3mm}
\begin{tabularx}{\linewidth}{
  >{\raggedright\arraybackslash}X
  >{\raggedright\arraybackslash}p{2.2cm}
  >{\raggedright\arraybackslash}p{1.6cm}
  >{\raggedleft\arraybackslash}p{1.5cm}
  >{\centering\arraybackslash}p{1.0cm}}
\toprule
\rowcolor{headerblue}
\textcolor{white}{\textbf{Dataset}} &
\textcolor{white}{\textbf{Venue}} &
\textcolor{white}{\textbf{Mechanism}} &
\textcolor{white}{\textbf{Size}} &
\textcolor{white}{\textbf{Avail.}} \\
\midrule

\rowcolor{groupA}
\multicolumn{5}{p{\linewidth}}{\textit{\textcolor{headerblue}{\textbf{StaVT: Meme} \quad|\quad \textbf{Goal:} Humor \& Entertainment, Satire \& Social Critique}}} \\
\rowcolor{groupA} \textbf{MemeReaCon}~\cite{zhao2025memereacon} & EMNLP'25 & Multi & 1{,}565 & \textcolor{naGray}{\textit{N/A}} \\
\rowcolor{groupA} \textbf{MEMESAFETY-BENCH}~\cite{lee2025vision} & EMNLP'25 & Multi & 50{,}430 & \href{https://huggingface.co/datasets/oneonlee/Meme-Safety-Bench}{Link} \\
\rowcolor{groupA} \textbf{MemeMind}~\cite{gu2025mememind} & Arxiv'25 & Multi & 43{,}223 & \textcolor{naGray}{\textit{N/A}} \\
\rowcolor{groupA} \textbf{MemeQA}~\cite{nguyen2025memeqa} & ACL'25 & Multi & 9K & \href{https://github.com/npnkhoi/memeqa}{Link} \\
\rowcolor{groupA} \textbf{SEMANTICMEMES}~\cite{zhou2024social} & NAACL'24 & Multi & 3.8M & \href{https://github.com/naitian/social-memeing}{Link} \\
\rowcolor{groupA} \textbf{MMD}~\cite{khan2024hope} & EMNLP-Fdgs'24 & Multi & 13{,}494 & \href{https://github.com/anas2908/MeSum}{Link} \\
\rowcolor{groupA} \textbf{MultiBully-Ex}~\cite{jha2024meme} & EACL'24 & Multi & 5{,}854 & \href{https://github.com/Jhaprince/MemeExplanation}{Link} \\
\rowcolor{groupA} \textbf{Oogiri-GO}~\cite{zhong2024let} & CVPR'24 & Multi & 130K & \href{https://huggingface.co/datasets/zhongshsh/CLoT-Oogiri-GO}{Link} \\
\rowcolor{groupA} \textbf{MemeMQACorpus}~\cite{agarwal2024mememqa} & ACL-Fdgs'24 & Multi & 1{,}880 & \textcolor{naGray}{\textit{N/A}} \\
\rowcolor{groupA} \textbf{ICMM}~\cite{jha2024memeguard} & ACL'24 & Multi & 1K & \href{https://github.com/Jhaprince/MemeGuard}{Link} \\
\rowcolor{groupA} \textbf{OxfordTVG-HIC}~\cite{li2023oxfordtvg} & ICCV'23 & Multi & 2.9M & \href{https://drive.google.com/drive/folders/1BDuUcMeaWrFD8TwgHLhFPkuAwmoHaVNQ}{Link} \\
\rowcolor{groupA} \textbf{MEMECAP}~\cite{hwang2023memecap} & EMNLP'23 & Multi & 6{,}300 & \href{https://github.com/eujhwang/meme-cap}{Link} \\
\rowcolor{groupA} \textbf{MCC}~\cite{sharma2023memex} & ACL'23 & Multi & 3.4K & \href{https://github.com/LCS2-IIITD/MEMEX_Meme_Evidence}{Link} \\
\rowcolor{groupA} \textbf{HatReD}~\cite{hee2023hatred} & IJCAI'24 & Multi & 3{,}228 & \href{https://github.com/Social-AI-Studio/HatReD}{Link} \\
\rowcolor{groupA} \textbf{ExHVV}~\cite{sharma2023you} & AAAI'22 & Multi & 3K & \href{https://github.com/LCS2-IIITD/LUMEN-Explaining-Memes}{Link} \\

\midrule
\rowcolor{groupB}
\multicolumn{5}{p{\linewidth}}{\textit{\textcolor{headerblue}{\textbf{StaVT: Cartoon} \quad|\quad \textbf{Goal:} Humor \& Entertainment}}} \\
\rowcolor{groupB} \textbf{HumorBench}~\cite{narad2025llms} & Arxiv'25 & Multi & 300 & \textcolor{naGray}{\textit{N/A}} \\
\rowcolor{groupB} \textbf{Humor in AI}~\cite{zhang2024humor} & NeurIPS'24 & Incongruity & 2.2M & \href{https://huggingface.co/datasets/yguooo/newyorker_caption_ranking}{Link} \\
\rowcolor{groupB} \textbf{Do Androids Laugh at Electric Sheep?}~\cite{hessel2023androids} & ACL'23 & Multi & 24{,}048 & \textcolor{naGray}{\textit{N/A}} \\

\midrule
\rowcolor{groupC}
\multicolumn{5}{p{\linewidth}}{\textit{\textcolor{headerblue}{\textbf{StaVT: Social Media Image} \quad|\quad \textbf{Goal:} Humor \& Entertainment, Satire \& Social Critique, Emotion \& Aesthetic Experience}}} \\
\rowcolor{groupC} \textbf{V-FLUTE}~\cite{saakyan2025understanding} & NAACL'25 & Multi & 6{,}027 & \href{https://huggingface.co/datasets/ColumbiaNLP/V-FLUTE}{Link} \\
\rowcolor{groupC} \textbf{SoMeLVLM}~\cite{zhang2024somelvlm} & ACL'24 & Multi & 653.8K & \href{https://huggingface.co/datasets/Lishi0905/SoMeData}{Link} \\

\midrule
\rowcolor{groupD}
\multicolumn{5}{p{\linewidth}}{\textit{\textcolor{headerblue}{\textbf{StaVT: Humorous Image} \quad|\quad \textbf{Goal:} Humor \& Entertainment}}} \\
\rowcolor{groupD} \textbf{HumorDB}~\cite{jain2025humordb} & ICCV'25 & Multi & 3{,}542 & \href{https://github.com/kreimanlab/HumorDB}{Link} \\
\rowcolor{groupD} \textbf{VisualPun\_UNPIE}~\cite{chung2024can} & EMNLP'24 & Incongruity & 1K & \href{https://huggingface.co/datasets/jiwan-chung/VisualPun_UNPIE}{Link} \\

\midrule
\rowcolor{groupA}
\multicolumn{5}{p{\linewidth}}{\textit{\textcolor{headerblue}{\textbf{SeqVN: Comic Strip} \quad|\quad \textbf{Goal:} Humor \& Entertainment}}} \\
\rowcolor{groupA} \textbf{MangaUB}~\cite{ikuta2025mangaub} & IEEE MM'25 & Narrative & 18{,}179 & \href{http://www.manga109.org/en/mangaub.html}{Link} \\
\rowcolor{groupA} \textbf{AI4VA-FG}~\cite{chen2025zooming} & Arxiv'25 & Narrative & 16{,}264 & \textcolor{naGray}{\textit{N/A}} \\
\rowcolor{groupA} \textbf{PixelHumor}~\cite{ryan2025humor} & EMNLP-Fdgs'25 & Multi & 2.8K & \href{https://github.com/Social-AI-Studio/PixelHumor}{Link} \\
\rowcolor{groupA} \textbf{YesBut-v2}~\cite{liang2025yes} & Arxiv'25 & Multi & 1{,}262 & \href{https://huggingface.co/datasets/zhehuderek/YESBUT_Benchmark}{Link} \\
\rowcolor{groupA} \textbf{YesBut}~\cite{hu2024cracking} & NeurIPS'24 & Multi & 348 & \href{https://huggingface.co/datasets/zhehuderek/YESBUT_Benchmark}{Link} \\
\rowcolor{groupA} \textbf{YesBut (synth.\ 3D stick)}~\cite{nandy2024yesbut} & EMNLP'24 & Multi & 2{,}547 (syn.) & \href{https://huggingface.co/datasets/bansalaman18/yesbut}{Link} \\
\rowcolor{groupA} \textbf{ComicsPAP}~\cite{vivoli2025comicspap} & Arxiv'25 & Narrative & 103{,}933 & \href{https://huggingface.co/datasets/VLR-CVC/ComicsPAP}{Link} \\

\midrule
\rowcolor{groupB}
\multicolumn{5}{p{\linewidth}}{\textit{\textcolor{headerblue}{\textbf{SeqVN: Meme} \quad|\quad \textbf{Goal:} Humor \& Entertainment}}} \\
\rowcolor{groupB} \textbf{XMeCap}~\cite{chen2024xmecap} & ACM MM'24 & Multi & 12{,}320 & \textcolor{naGray}{\textit{N/A}} \\

\bottomrule
\end{tabularx}
\label{tab:datasets3}
\end{table*}

\section{Task Specific Models Before MLLM Era}
\label{before_mllm}

Prior to MLLMs, multimodal visual humor understanding was dominated by task-specific discriminative architectures tightly coupled with individual tasks such as sarcasm, humor, or metaphor detection.

Early models focused on effective fusion mechanisms for heterogeneous features. \citet{schifanella2016detecting} first incorporated visual cues into sarcasm detection via separate encoders and concatenation, while \citet{cai2019multi} showed hierarchical fusion of text, images, and attributes better captures cross-modal interactions. \citet{zhang2021multi} introduced contrastive attention to explicitly model inter-modal incongruity for finer detection of cross-modal discrepancies. These approaches improved reasoning on specific tasks, but they also introduced stronger task assumptions and often depended on curated knowledge sources or structures.

Beyond fusion design, later work emphasized implicit knowledge and commonsense. HKT~\cite{hasan2021humor} injected humor-related knowledge into Transformer architecture for deeper incongruity modeling. \citet{lee2021disentangling} and \citet{liang2022multi} leveraged object-level visual representations to construct richer contextual embeddings and cross-modal graphs, facilitating localized reasoning over visual–textual conflicts. \citet{liu2022towards} incorporated external commonsense and semantic knowledge into hierarchical congruity modeling, improving implicit intent interpretation.

Overall, non-MLLM approaches made important progress by improving fusion design, structured representations, and knowledge injection. However, their gains were typically task-dependent, and their scalability and cross-domain generalization remained limited, as also noted in prior surveys~\cite{sharma2022detecting,farabi2024survey}. This limitation motivates the later shift toward more general multimodal foundation-model paradigms.

\end{document}